\documentclass[runningheads]{llncs}

\usepackage{graphicx}
\usepackage{color}
\usepackage{algorithmic}
\usepackage{algorithm}
\usepackage{multirow}
\usepackage{comment}
\usepackage[hidelinks]{hyperref}

\begin{document}

\title{Exploring Interpretability\\ for Predictive Process Analytics}
\author{Renuka Sindhgatta \and 
Chun Ouyang \and 
Catarina Moreira} 
%
\institute{Queensland University of Technology, Brisbane, Australia\\ 
\email{\{renuka.sr,c.ouyang,catarina.pintomoreira\}@qut.edu.au}}

\maketitle 

\vspace*{-\baselineskip}

\begin{abstract}
Modern predictive analytics underpinned by machine learning techniques has become a key enabler to the automation of data-driven decision making. In the context of business process management, predictive analytics has been applied to making predictions about the future state of an ongoing business process instance, for example, when will the process instance complete and what will be the outcome upon completion. Machine learning models can be trained on event log data recording historical process execution to build the underlying predictive models. Multiple techniques have been proposed so far which encode the information available in an event log and construct input features required to train a predictive model. While accuracy has been a dominant criterion in the choice of various techniques, 
these techniques are often applied as a black-box in building predictive models. In this paper, we derive explanations using interpretable machine learning techniques to compare and contrast the suitability of multiple predictive models of high accuracy. The explanations allow us to gain an understanding of the underlying reasons for a prediction and highlight scenarios where accuracy alone may not be sufficient in assessing the suitability of techniques used to encode event log data to features used by a predictive model. Findings from this study motivate the need and importance to incorporate interpretability in predictive process analytics. 
\end{abstract}

\section{Introduction}
\label{sec:intro}

Modern predictive analytics underpinned by machine learning techniques has become a key enabler to the automation of data-driven decision making. In the context of business process management, predictive process analytics is a relatively new discipline that aims at predicting future observations of a business process by learning from event log data that capture the process execution history. A vast majority of work over the past decade has used supervised machine learning algorithms to construct predictive models for predicting outcomes of a business process instance (or a case)~\cite{teinemaa2019}, the next activity in a case~\cite{fettke2017}, or the remaining time for a case to complete~\cite{verenich2019}. Evaluation of a predictive model has so far been assessed in terms of the quality of the learning model and thus evaluated using conventional metrics in machine learning (such as accuracy, precision, recall, F1 score). 

As an important branch of state-of-the-art data analytics, predictive process analytics is also faced with a challenge in regard to the lack of explanation to the reasoning and outcome of its predictive models. While more and more complex machine learning techniques (and more recently, deep learning techniques) are used to build advanced predictive capabilities in process analytics, they are often applied and recognised as a `black-box'. 

The recent body of literature in machine learning has emphasised the need to understand and trust the predictions (e.g.,~\cite{lakkaraju2019,rudin2019}). This has led to an increasing interest in the research community on \textit{interpretable} machine learning~\cite{guidotti2018}. 
``To \textit{interpret} means to give or provide the meaning or to explain and present in understandable terms''~\cite{guidotti2018}. Having an interpretable or explainable model is a necessary step towards obtaining a good level of understanding about the rationale of the underlying `black-box' machinery. 

In this paper, we derive interpretation of the predictive models trained with various input features representations of the event logs. We review the techniques that have been evaluated in the benchmark studies on business process monitoring benchmarks for predicting process outcomes and remaining time. By applying interpretable machine learning techniques to two existing benchmarks~\cite{teinemaa2019,verenich2019}, we derive global explanations that present the behaviour of the entire predictive model as well as local explanations describing a particular prediction. These explanations are useful for reviewing the suitability of a model when predicting process behaviour as well as for understanding the importance and relevance of certain features used for prediction. Findings drawn from this work are expected to motivate the need to incorporate interpretability in predictive process analytics. To the best of our knowledge, the closest study to this work is an illustration of the potential of explainable models for a manufacturing business process~\cite{rehse2019}.

The rest of the paper is structured as follows. Sect.~\ref{sec:background} provides necessary background information on predictive process monitoring benchmarks and interpretability of predictive models. Sect.~3 presents a detailed review of two existing predictive process monitoring benchmarks by interpreting predictive models in several methods. Sect.~4 discusses and summarizes several findings drawn from our analyses. Finally, Sect.~5 concludes the paper. 
\section{Background} 
\label{sec:background}

\subsection{Predictive Process Monitoring Benchmarks}
\label{subsec:benchmarks}

In this work, two studies~\cite{teinemaa2019,verenich2019} that evaluate various techniques used in the context of predictive process monitoring are considered. Existing studies predicting future behaviour of business process mainly apply supervised machine learning algorithms. A supervised learning algorithm uses a corpus of input, output pairs (or training data) to learn a hypothesis that can predict the output for a new or unseen input (test data). The input and output are derived from the event logs. The output in the two benchmarks are: 1) an outcome of a case defined by using a labelling function, and 2) the remaining time for a case to complete. 

\subsubsection{Predictive Process Monitoring Approach.} 

\begin{figure}[h!]
\vspace*{-.75\baselineskip}
   \resizebox{\columnwidth}{!}{\centering\includegraphics{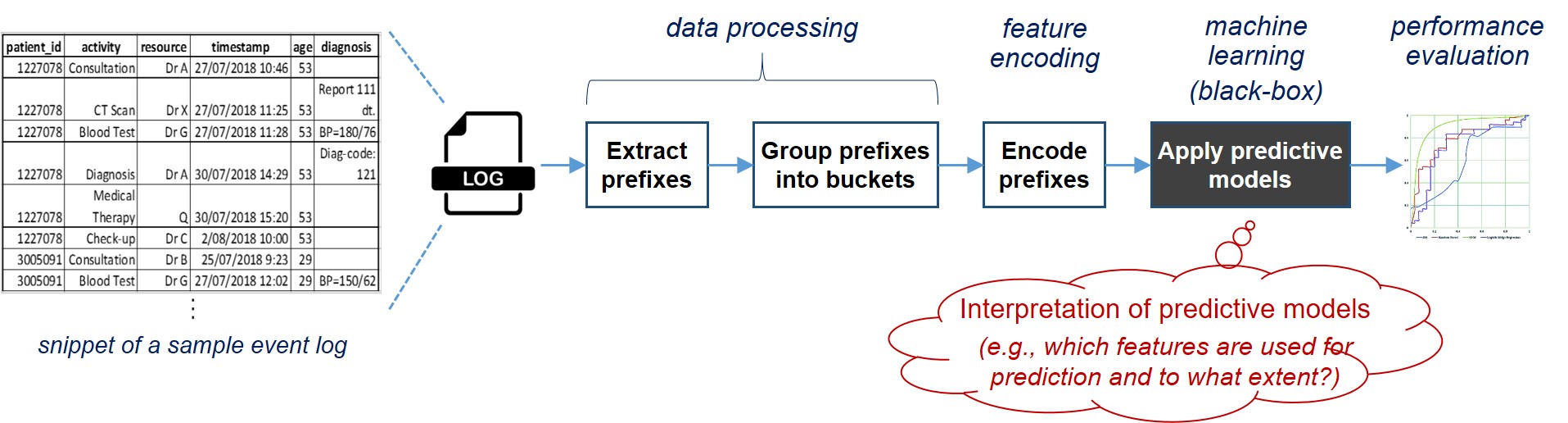}}
		\vspace*{-1.25\baselineskip}
    \caption{Overall approach of predictive analytics for process monitoring}
\vspace*{-1.5\baselineskip}
\label{fig:prediction_workflow}
\end{figure}

An overall approach of predictive analytics used in the context of process monitoring is illustrated in Fig.~\ref{fig:prediction_workflow}. More specifically, a \textit{trace}~$\sigma$ is a sequence of \textit{events} of the same case. 
During the training phase, \textit{prefixes} are generated for each trace. A prefix function $\mathit{prefix}(\sigma,l)$ takes as input a trace~$\sigma$ and a prefix of length~$l$, and returns the first~$l$ events of the trace. The prefixes are grouped into \textit{buckets} based on their similarities (such as length, process states, or events) using a \textit{bucketing mechanism}. The prefixes in each bucket are encoded as feature vectors following an \textit{encoding mechanism}. The buckets of feature vectors are then used to train a predictive model underpinned by a (machine) \textit{learning algorithm}. Since the future state for each case is known from the training data, pairs of the encoded prefix and the future state are used to train a predictive model for each bucket. 

 
\paragraph{Evaluation Measures.}

Existing predictive process monitoring methods are built on different combinations of bucketing mechanisms, encoding mechanisms, and learning algorithms. The two predictive process monitoring benchmarks~\cite{teinemaa2019,verenich2019} evaluate these methods using the following quality measures: 
\begin{itemize}
	\item \textit{Accuracy}: for outcome-oriented prediction, this is measured by the \textit{area under the ROC curve} (AUC) metric; and for remaining time prediction, this is measured by the \textit{mean absolute error} (MAE) metric. Higher the AUC, better the model is at predicting and distinguishing the outcomes. MAE for remaining time prediction is the absolute difference between the actual remaining time and predicted remaining time and a lower MAE indicates better performance of the predictive model.
	\item \textit{Earliness}: in both predictions, this is defined as the smallest prefix length with the desired level of accuracy. 
\end{itemize}

 \subsection{Interpretability of Predictive Models}
 \label{subsec:predictivemodels}

In recent years, the topic of \textit{interpretable} or \textit{explainable} machine learning has gained attention. To avoid ambiguity, we apply the terms \textit{interpretability} and \textit{explainability} as discussed in the machine learning literature~\cite{guidotti2018,lipton2018}. Model interpretability can be addressed by having \textit{intepretable models} and/or providing \textit{post hoc interpretations}. An \textit{interpretable model}~\cite{lipton2018}, is able to provide transparency at the level of entire model (\textit{simulatability}), the level of individual components (\textit{decomposability}), and the level of the learning algorithm (\textit{algorithmic transparency}). For example, both linear regression models and decision tree models are interpretable models, while neural network models are complex and hard to follow and hence have low transparency.

Another distinct approach to address model interpretability is via \textit{post hoc interpretation}. Here, explanations and visualisations are extracted from a learned model, that is, \textit{after} the model has been trained, and hence are \textit{model agnostic}. Existing techniques can be divided into two categories: partial dependence models and surrogate models. 
More specially, \textit{surrogate models} use the input data and a black box model (i.e., a trained machine learning model) and emulate the black box model. In other words, they are approximation models that use interpretable models 
to approximate the predictions of a black box model, enabling a decision-maker to draw conclusions and interpretations about the black box~\cite{Molnar18}. 
Interpretable machine learning algorithms, such as linear regression and decision trees are used to learn a function using the predictions of the black box model. 
This means that this regression or decision tree will learn both well classified examples and misclassified ones. 
Measures such as mean squared error are used to assess how close the predictions of the surrogate model approximate the black box.
As a result, the explanations derived from the surrogate model reflect a local and linear representation of the black box model. 
In more detail, the general algorithm for surrogate models is presented in Algorithm~1 (from~\cite{Molnar18}). 
\vspace*{-.5\baselineskip}


\begin{algorithm} [h!]
\label{alg:surrogate}
\caption{General algorithm for surrogate models~\cite{Molnar18}}
\label{alg:index}
\begin{algorithmic}[1]
\REQUIRE Dataset~$\mathcal{X}$ used to train black box, Prediction model $\mathcal{M}$\\
\ENSURE  Interpretable Surrogate model I \\ 
\vspace*{-.5\baselineskip}
~~\\
\STATE Get the predictions for the selected $\mathcal{X}$, using the black box model~$\mathcal{M}$\\
\STATE Select an interpretable model: linear model, regression tree, ... \\
\STATE Train interpretable model on $\mathcal{X}$, obtaining model $\mathcal{I}$ \\
\STATE Get predictions of interpretable model $\mathcal{I}$ for $\mathcal{X}$ \\
\STATE Measure the approximation of interpretable model $\mathcal{I}$ with the black box model~$\mathcal{M}$\\
\RETURN Interpretation of $\mathcal{I}$

\end{algorithmic}
\end{algorithm}
\vspace*{-1\baselineskip}

\section{Interpreting Predictive Models for Process Monitoring}
\label{sec:reviewbenchmarks}

In this section, we aim to interpret and gain deeper insights into the predictive models used for predicting process outcome and remaining time. In the context of the two existing process monitoring benchmarks, we first derive global and local explanations of selected predictive models using model interpretation techniques, and then conduct several analyses of the derived interpretations to draw interesting findings about these predictive models. Our detailed analysis, as well as the source code, is available at \url{https://git.io/Je186} (for interpreting process outcome prediction) and \url{https://git.io/Je1XZ} (for interpreting process remaining time prediction).

\subsection{Design and Configuration}
\label{subsec:approach} 

Combinations of various bucketing, encoding, and supervised learning algorithms have been evaluated for predicting process outcome~\cite{teinemaa2019} and remaining time~\cite{verenich2019}, respectively. In this study, we decide to choose the following techniques, because the methods built upon a combination of these techniques have better performance (i.e. high AUC values or low MAE values) as compared to others according to the benchmark evaluation~\cite{teinemaa2019,verenich2019}. 

\paragraph{Bucketing techniques:}  
    i) \textit{single bucket}, where all prefixes of traces are considered in a single bucket, and a single classifier is trained; and 
    ii) \textit{prefix length bucket}, where each bucket contains partial traces of a specific length, and one classifier is trained for each possible prefix length.
							
\paragraph{Encoding techniques:}  
	i) \textit{aggregation encoding}, where the trace in each bucket is transformed by considering (only) the frequencies of event attributes such as activity, resource and computing four features for the numeric event attributes (max, mean, sum and standard deviation), and note that this way the order of the events in a trace is ignored; 
	ii) \textit{index encoding}, where each event attribute (e.g., activity, originator/resource) of an executed event can be represented as a feature, and this way the order of a trace in each bucket can be maintained by encoding each event in the trace at a given index; and 
	iii) \textit{static encoding}, where the trace (or case) attributes that remain the same through out the trace is added as a feature as-is. The categorical attributes are one-hot encoded, where each categorical attribute value is represented as a feature that takes a binary value of 0 or 1. The encoding methods are summarised in Table~\ref{tab:encoding}.
	
\begin{table}[h!]
\vspace*{-.75\baselineskip}
\caption{Overview of the selected encoding methods}
\vspace*{-.5\baselineskip}
\scriptsize
\centering
\begin{tabular}{|p{2cm}|p{3cm}|p{3.3cm}|p{3.3cm}|}
\hline
\multirow{2}{*}{\bf Encoding}       & \multirow{2}{*}{\bf Attributes}       & \multicolumn{2}{p{3.3cm}|}{\bf Feature Extraction}   \\ 
\cline{3-4}                         &                                       & \it Numeric               & \it Categorical                   \\ 
\hline
\it \textbf{Static}                 & Case                                  & as-is                     & one-hot                       \\ 
\hline
\it \textbf{Aggregation}            & Events (unordered)                    & min, max, mean,           & frequencies, or               \\ 
                                    &                                       & std. deviation            & number of occurrences             \\ 
\hline
\it \textbf{Index}                  & Event (ordered)                       & as-is for each index      & one-hot for each index        \\ 
\hline
\end{tabular}
\vspace*{-1.25\baselineskip}
\label{tab:encoding}
\end{table}

\paragraph{Machine learning algorithms:} 
    We choose \textit{Gradient boosted trees}~\cite{friedman2001} (specifically XGBoost), which is used in both benchmarks and outperformed the other machine learning techniques (e.g., random forest, support vector machines, logistic regression). 
    Note that this study focuses on machine learning algorithms and hence Long Short Term Memory (LSTM)~\cite{verenich2019}, which is used for predicting the remaining time of running cases, is not considered. 
    
\paragraph{Interpretation techniques:} 
    To make sure that the output and performance of a predictive method being studied remain intact, we apply post-hoc interpretation to derive explanations for the predictive methods built upon the above techniques and algorithms in both benchmarks. We choose a representative local surrogate method, known as \textit{Local Interpretable Model-Agnostic Explanations} (LIME)~\cite{Ribeiro16}, which can explain the predictions of any classification or regression algorithm, by approximating it locally with a linear interpretable model. We use LIME to generate \textit{local} explanations useful in interpreting the prediction for a particular trace. In addition, we also conduct permutation feature importance measurement supported by gradient boosted trees to gain certain \textit{global} explanations about a predictive method. More specifically, the feature importance value generated by XGBoost is used to explain the impact of different features to the overall predictions made by a given predictive method.  

\subsection{Datasets and Notations}
\label{subsec:data} 

We present the results on three real-life event logs that are representative of our analysis. These event logs are from the Business Process Intelligence Challenge (BPIC~2011\footnote{\url{https://data.4tu.nl/repository/uuid:d9769f3d-0ab0-4fb8-803b-0d1120ffcf54}},  BPIC~2012\footnote{\url{https://data.4tu.nl/repository/uuid:3926db30-f712-4394-aebc-75976070e91f}} and BPIC~2015\footnote{\url{https://data.4tu.nl/repository/uuid:31a308ef-c844-48da-948c-305d167a0ec1}}), and were used for performance evaluation of predictive methods in both process monitoring benchmarks~\cite{teinemaa2019,verenich2019}. 
Below are brief descriptions of each of the logs as well as certain notations to be used in the interpretations and analysis. 

\paragraph{BPIC 2011:} The event log contains cases from the Gynaecology department of a Dutch hospital. For interpreting the outcome prediction, the outcome labelling function based on the occurrence of activity \textit{``histological examination - big resectiep''} is used (i.e., \textit{bpic2011\_4}). In the log preprocessing, the trace for each case is cut exactly before this event occurs. For interpreting the  remaining time prediction, the log is used as-is without any truncation (\textit{bpic2011}). 

\paragraph{BPIC 2012:} The event log contains the traces of a loan application process at a Dutch financial institution. For the outcome prediction, each trace in the log is labelled as \textit{``accepted''}, \textit{``declined''}, or \textit{``cancelled''} (based on whether the trace contains the occurrence of activity \texttt{O\_ACCEPTED}, \texttt{O\_DECLINED}, or \texttt{O\_CANCELLED}). One of the three logs, which concerns loan acceptance, is considered for generating model explanations (\textit{bpic2012\_1}). For the remaining time prediction, three logs are generated depicting loan application, loan offers and loan processing by human workers, respectively. Explanations are presented for the model trained on the loan offers (\textit{bpic2012o}).

\paragraph{BPIC 2015:} 
There are five event logs recording the traces of a permit application process at five Dutch municipalities, respectively. 
For the outcome prediction, the rule stating every occurrence of activity \texttt{01\_HOOFD\_020} is eventually followed by activity \texttt{08\_AWB45\_020\_1} is used to label the trace as \textit{positive}. Explanations are derived for one of the municipalities (\textit{bpic2015\_5}).

\paragraph{Notations:} 
The feature representations in the graphical figures in Sect.~\ref{subsec:analysis} are to be read as follows: 
\begin{itemize}
    \item \texttt{agg\_\_[Activity]|[Resource]} represents the frequency of an activity executed by a resource) in a trace via aggregation encoding.
    \item \texttt{index\_\_[Activity]|[Resource]\_[idx]\_[name]} represents the index (\texttt{idx}) at which an activity or a resource of a given \texttt{name} occurs via index encoding.
    \item \texttt{static\_\_[attribute]} represents the case attribute for a trace (of which the value does not change during the execution of the trace).
    \item \texttt{agg\_max/mean/std\_[Attribute]} represents the descriptive statistics (maximum, mean, standard deviation) of numeric event attributes via aggregation encoding. 
\end{itemize}

\subsection{Interpretations and Analysis}
\label{subsec:analysis} 




\subsubsection{Analysis~1: Single bucket vs. prefix-length bucket.} 

\begin{figure}[b!]
    \resizebox{\textwidth}{!}{\centering\includegraphics{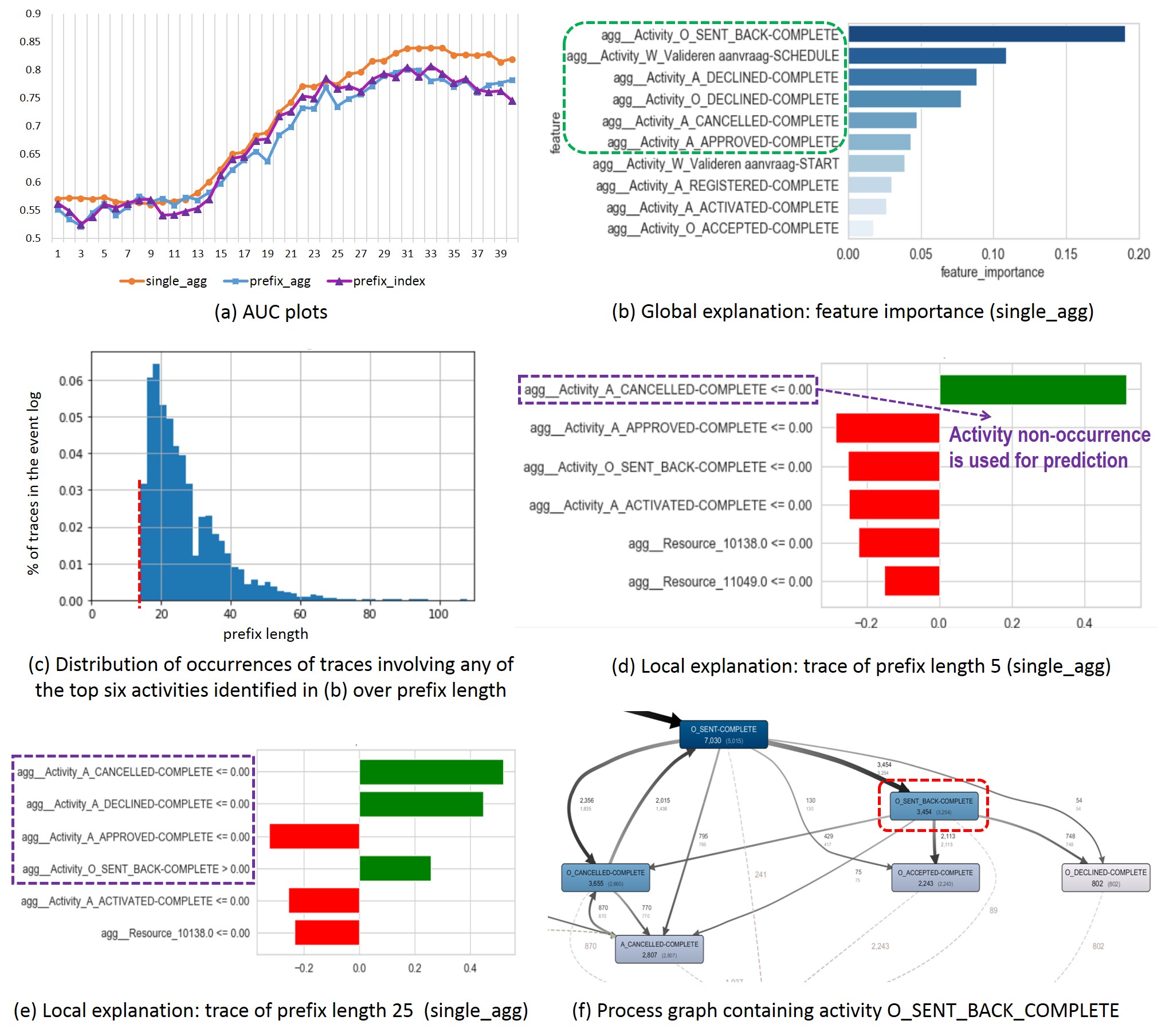}}
    \caption{Interpreting outcome prediction using \textit{bpic2012\_1} event log: 
    (a) AUC plots of the predictive methods built on \textit{single\_agg}, \textit{prefix\_agg} and \textit{prefix\_index} with XGBoost~\cite{teinemaa2019}; 
    (b)-(e) for \textit{single\_agg} method, where:  
    (b) global explanation using feature importance values, 
    (c) distribution of certain trace occurrences over prefix length used for prediction, and 
    (d) \& (e) local explanations for traces of prefix lengths $5$ \& $25$; 
    and finally (f) snapshot of discovered process model containing activity \texttt{O\_SENT\_BACK\_COMPLETE}.}
\label{fig:bpic2012_agg_single_prefix}
\end{figure}

This is to compare the two different bucketing techniques in terms of their impact on predictions. The following two combinations are applied to \textit{bpic2012\_1} event log for outcome prediction: i) single bucket and aggregation encoding (\textit{single\_agg}) and ii) prefix bucket and aggregation encoding (\textit{prefix\_agg}). 
According to the evaluation results in~\cite{teinemaa2019}, overall the \textit{single\_agg} method has better AUC and earliness values compared to the \textit{prefix\_agg} method (refer to the AUC plots in Fig.~\ref{fig:bpic2012_agg_single_prefix}(a)). 

We analyse \textit{single\_agg} method where a single classifier XGBoost is trained for a single bucket containing all traces of all prefix lengths. Starting with global explanations, Fig.~\ref{fig:bpic2012_agg_single_prefix}(b) shows the importance of the features used by the model trained using \textit{single\_agg} with XGBoost. The model used the occurrences of activities \texttt{O\_SENT\_BACK}, \texttt{W\_Validate Request}, \texttt{A\_DECLINED}, \texttt{O\_DECLINED}, \texttt{A\_CANCELLED} and \texttt{A\_APPROVED} as the top six important features. A statistical analysis of the event log data reveals that the above six activities occur in the traces at a minimum prefix length of $14$ (refer to Fig.~\ref{fig:bpic2012_agg_single_prefix}(c)). This means that when predicting the outcome of running traces with prefix lengths lower than $14$,  \textit{a \textit{single\_agg} method will likely use zero occurrences of those six activities to make the prediction}. Hence, based on global explanations, it can be inferred that for traces of lower prefix lengths ($<14$), the features considered important for prediction will have zero values when using a predictive method of \textit{single\_agg} with XGBoost. 

We then generate \textit{local explanations} for two randomly chosen traces for which the model correctly predicted the positive outcome (i.e., loan accepted). Fig.~\ref{fig:bpic2012_agg_single_prefix}(d) and~(e) shows local explanations for traces with prefix lengths of $5$ and $25$, respectively. At lower prefix length ($5$), while the model predicts accurately,  the non-occurrence of the activity \texttt{A\_CANCELLED} influences the prediction. Use of non-occurrence of an activity to make a prediction may not be a reliable feature to use. In contrast, we observe that for higher prefix length ($25$), the model uses the occurrence of activity \texttt{O\_SENT\_BACK} as an important feature when predicting the outcome, which is a reliable feature to use as the process model discovered from the event log reveals that a loan may be accepted after its occurrence (Fig.~\ref{fig:bpic2012_agg_single_prefix}(f)). 
Hence, based on local explanations, we can conclude that while a predictive method using \textit{a single bucket} may present higher accuracy and earliness, it would not be suitable for traces with lower prefix lengths. 

\begin{figure}[t!]
		\vspace*{.5\baselineskip}
		\resizebox{\textwidth}{!}{\centering\includegraphics{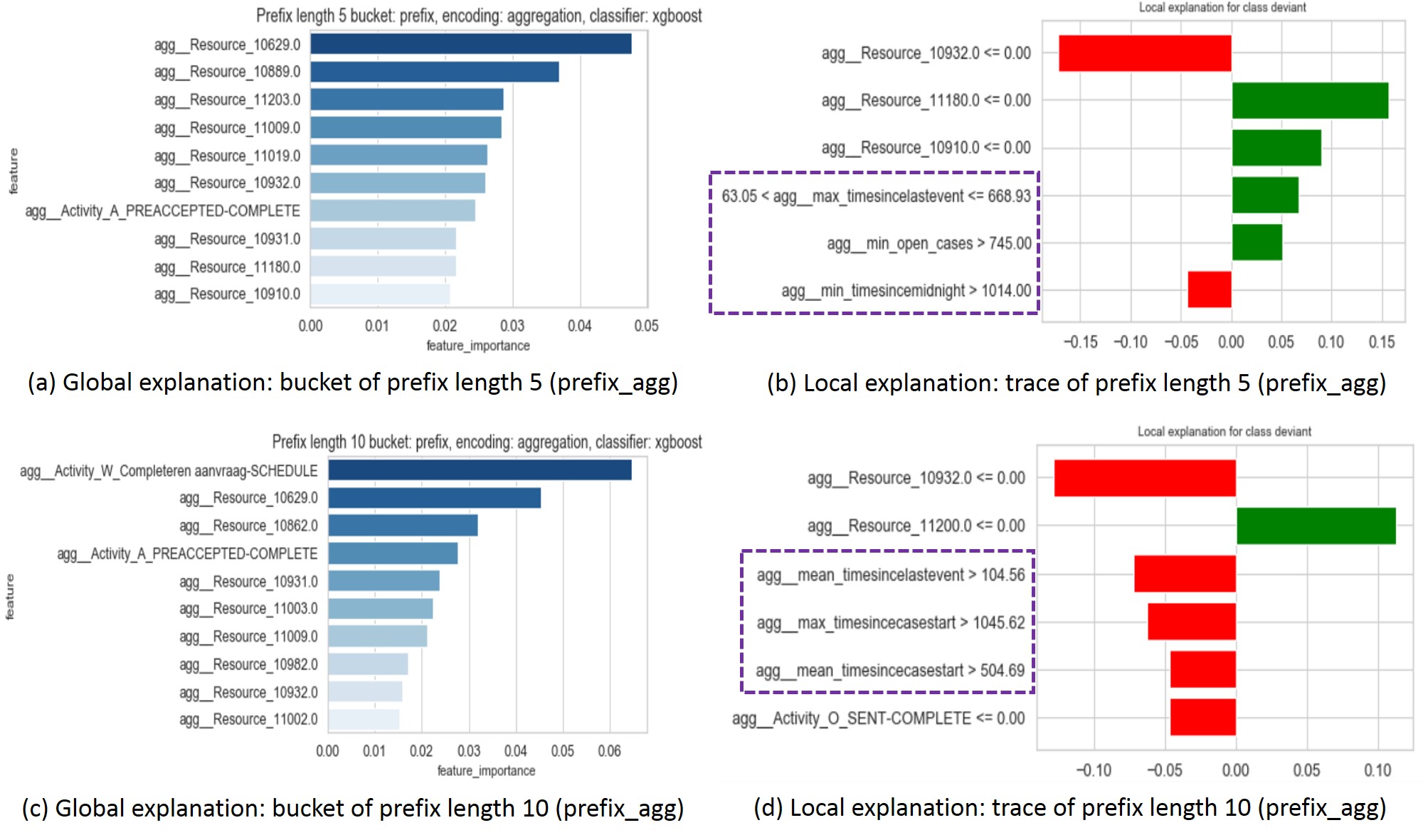}}
		\caption{Interpreting outcome prediction using \textit{bpic2012\_1} event log: (a)-(d) global and local explanations for \textit{prefix\_agg} method with prefix lengths $5$ and $10$.} 
\label{fig:bpic2012_prefix_agg}
\end{figure}

We further analyse \textit{prefix\_agg} method where a classifier XGBoost is trained for each bucket containing traces of certain, pre-defined prefix lengths. Global and local explanations for buckets containing prefixes of length $5$ and $10$ are shown in Fig.~\ref{fig:bpic2012_prefix_agg}. 
Global explanations indicate that the important features are resources executing the activities (Fig.~\ref{fig:bpic2012_prefix_agg}(a) and (c)). 
Local explanations provide some interesting insights: lower values of the features (such as \textit{the time since last event}, \textit{the time since the case started}) increase the likelihood of the positive outcome (Fig.~\ref{fig:bpic2012_prefix_agg}(b) and (d)). While a \textit{prefix\_agg} method has generally lower accuracy (refer to AUC plots in Fig.~\ref{fig:bpic2012_agg_single_prefix}(a)), it uses features that can be computed based on the activities executed in the running case of a given prefix length and hence would be better suited for traces with lower prefix lengths. 

\subsubsection{Analysis~2: Aggregation encoding vs. index encoding.} 

This is to compare the two different feature encoding techniques in terms of their impact on predictions. Based on the findings in Analysis~1, we use prefix-length bucket and apply the following two combinations with XGBoost models to \textit{bpic2012\_1} event log for outcome prediction: i) prefix bucket and aggregation encoding (\textit{prefix\_agg}) and ii) prefix bucket and index encoding (\textit{prefix\_index}). Both models are of similar accuracy according to the AUC plots shown in Fig.~\ref{fig:bpic2012_agg_single_prefix}(a). 

\begin{figure}[t!]
		\vspace*{.5\baselineskip}
		\resizebox{\textwidth}{!}{\centering\includegraphics{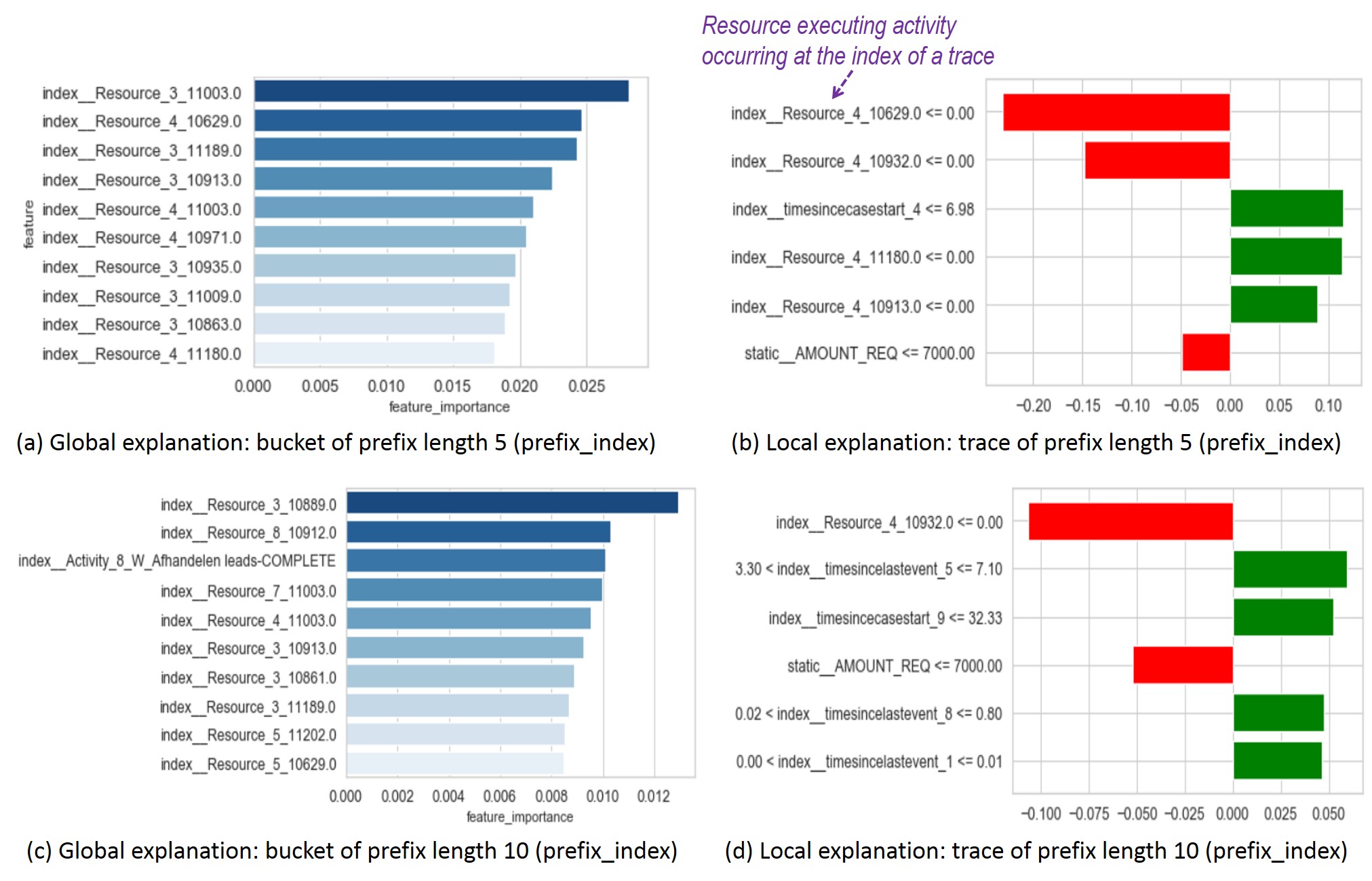}}
		\caption{Interpreting outcome prediction using \textit{bpic2012\_1} event log: (a)-(d) global and local explanations for \textit{prefix\_index} method with prefix lengths $5$ and $10$.}
\label{fig:bpic2012_prefix_index}
\end{figure}

Fig.~\ref{fig:bpic2012_prefix_index}(a) and~(c) illustrate the global explanations of the model trained at distinct prefix lengths ($5$ and $10$) using index encoding, which show that the model used activity or resource information about the prior events. The features used by the model are similar to the aggregation encoding for prefix-length buckets (refer to Fig.~\ref{fig:bpic2012_prefix_agg}(a) and~(c)). The models trained on lower prefix lengths (e.g., $5$) use information about the resources executing activities at certain indexes (temporally ordered). 

While both \textit{prefix\_agg} and \textit{prefix\_index} use information of the resources, the resource values used by the models are different. In this scenario, we are unable to conclude which model provides better explanations limited by our knowledge of the business process. However, since the index encoding uses attributes at each index, the number of features can be very high for processes that have large number of activities, resources and data attributes. The accuracy of the model could deteriorate in those situations as observed in the benchmark results presented in~\cite{teinemaa2019}. Hence, aggregation encoding would be considered more suitable when dealing with business processes with large number of activities, resources and data attributes.

\subsubsection{Analysis~3: One-hot data encoding.} 

In both process outcome and remaining time predictions, the aggregation and index encoding techniques further apply one-hot data encoding (or, one-hot encoding) to represent the activities, resources and other categorical data from the event log to feature vectors as input for machine learning models. The purpose of this analysis is to understand the impact of one-hot encoding on the predictive methods being studied. 

In principle, one-hot encoding increases the size of a dataset exponentially, because each attribute value of a feature becomes a new feature by itself with the possible value of $0$ or $1$. For instance, a feature~$F$ with three attributes values $f_1$, $f_2$ and $f_3$ will be represented as three new binary features. This data encoding method generates very sparse datasets, which impacts negatively both the performance metrics of the prediction model and the ability to generate interpretations. Below we discuss the findings that were obtained from analysing the explanations derived from applying \textit{single\_agg} with XGBoost model to process remaining time prediction using \textit{bpic2011} event log as an example. 

\begin{figure}[b!]
    \resizebox{\textwidth}{!}{\centering\includegraphics{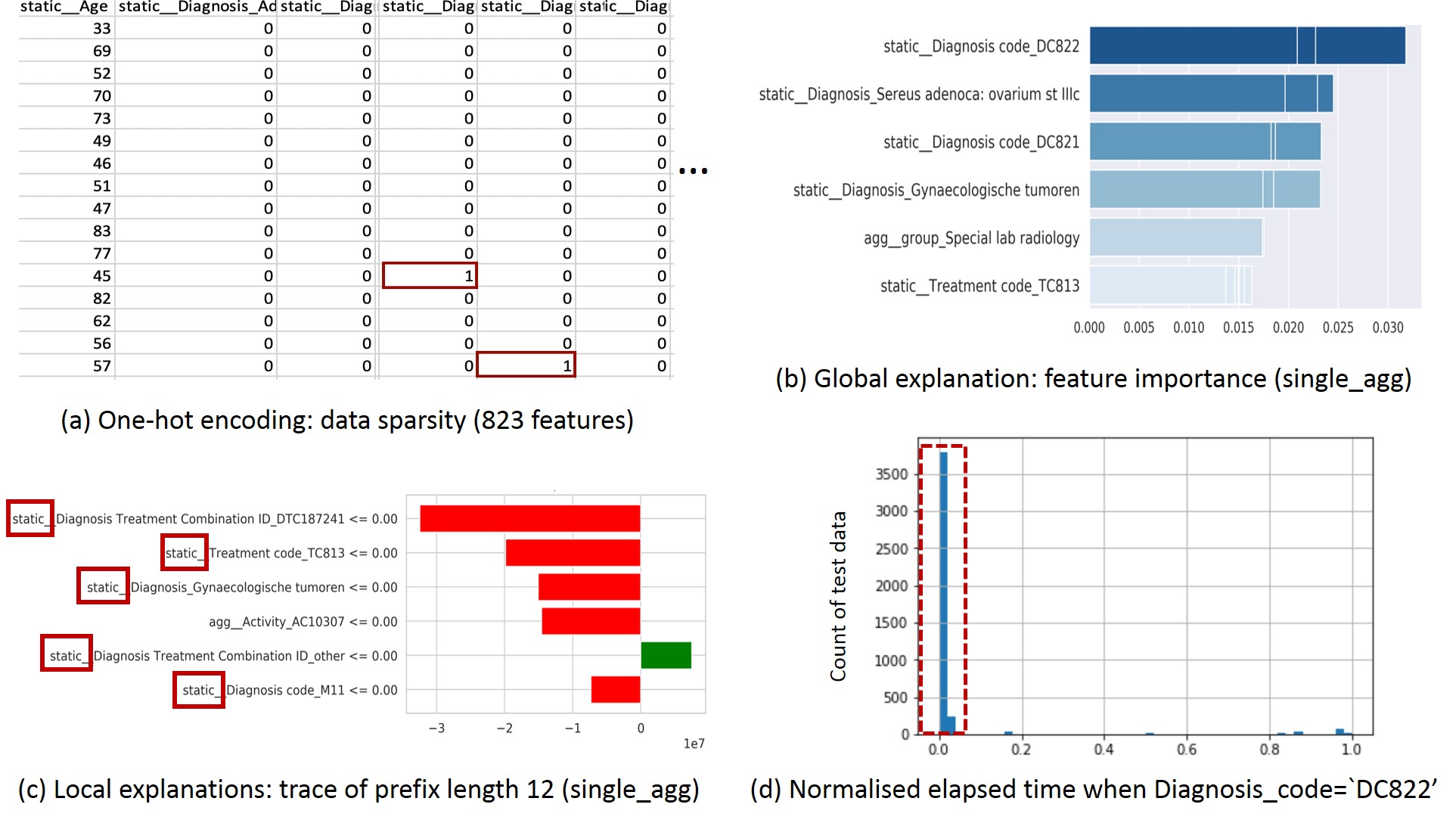}}
    \caption{Interpreting remaining time prediction for \textit{bpic2011} event log using \textit{single\_agg} with XGBoost: (a) data sparsity as result of one-hot encoding; (b) global explanation; (c) local explanation for trace of prefix length of $12$; and (d) normalised elapse time for a specific diagnosis code.}
\label{fig:bpic2011_tp}
\end{figure}

Fig.~\ref{fig:bpic2011_tp} depicts certain impact of one-hot encoding in the context of \textit{bpic2011} log. The original dataset increased from approximately 20 features to 823 features with this representation (see a snapshot of feature matrix in Fig.~\ref{fig:bpic2011_tp}(a)). Further, a majority of the local explanations can be represented as the follows: \textit{if feature~$X$ is absent (value $\leq 0$), then it influences the remaining time prediction}. When a dataset is so sparse, it is reasonable to expect such type of explanations. However, the question that may arise from this finding is: \textit{To what extent can this provide a meaningful understanding of why the predictive model made a certain prediction?}


\subsubsection{Analysis~4: Feature relevance.} 

The purpose of this analysis is to reason the relevance of the features identified important for process prediction. Feature importance in global explanations and feature impact to predictions of traces in local explanations are valuable inputs to interpreting feature relevance. To derive such interpretation a good understanding of the business process is often needed. Below, we discuss two examples of applying \textit{single\_agg} with XGBoost to remaining time prediction using \textit{bpic2011} and \textit{bpic2012o} event logs, respectively. 

For \textit{bpic2011} log, it can be observed that, from the local explanation shown in Fig.~\ref{fig:bpic2011_tp}(c), the predictive model relies on features such as \texttt{Diagnosis Treatment Combination ID}, \texttt{Treatment code} and \texttt{Diagnosis code} in order to determine the remaining time of the case. For a regression problem, like time prediction, this explanation indicates that the model is relying mostly on \textit{static} features, which are the features that do not change throughout the lifetime of a case. The usage of static features for regression suggests that the process execution does not rely on the executions of activities or cases (i.e., sequences of activities following different control-flow logics) and that the model uses attributes that do not change during the case execution when making a prediction. 
Another interesting observation is from the global explanation shown in Fig.~\ref{fig:bpic2011_tp}(b), which indicates that the most significant feature for remaining time prediction is \texttt{Diagnosis\_code = DC822}. By applying statistical analysis on the event log, as depicted in Fig.~\ref{fig:bpic2011_tp}(d), we discovered that 82\% of the events associated with this diagnosis code had the feature \texttt{elapsed time = 0}, which means that the corresponding activity starts and ends immediately. To this end, lack of relevant knowledge about the business process limits our ability to derive further insights about the relationship between these static diagnosis codes, with 0-valued elapsed times and its relation with the remaining time of a running case of the process.

\begin{figure}[h!]
		\vspace*{-.5\baselineskip}
    \resizebox{\textwidth}{!}{\includegraphics{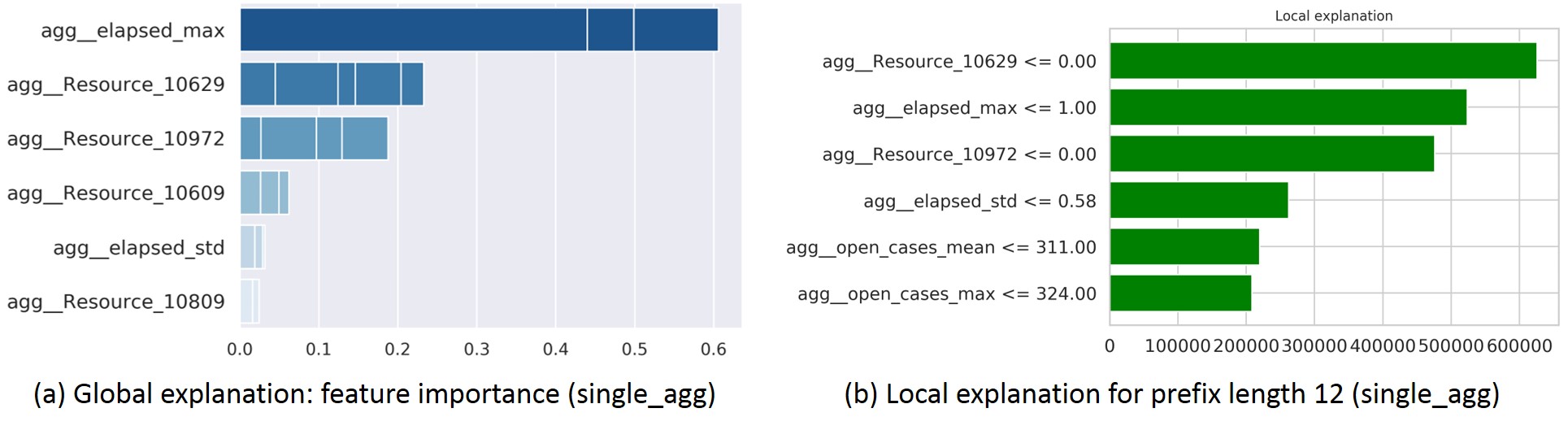}}
    \caption{Interpreting remaining time prediction for \textit{bpic2012o} event log using \textit{single\_agg} with XGBoost: (a) global explanation and (b) local explanation for prefix length $12$.}
\label{fig:bpic2012O_tp}
\end{figure}

For \textit{bpic2012o} log, the analysis leads to different observations. As shown in Fig.~\ref{fig:bpic2012O_tp}, the global explanation and the local explanation for trace of prefix length $12$ indicate the resource perspective of the business process are the important features used by the predictive model and have a positive impact on the remaining time prediction. The features identified as highly relevant may be case-related (such as \texttt{agg\_opencases}), resource-related (such as \texttt{agg\_resource}), or time-dependent (such as \texttt{agg\_elapsed time}). It is worth noting that the features like \texttt{agg\_opencases} and \texttt{agg\_elapsed time}, which have a positive impact on the performance of the predictive model, are not among the data attributes of the original event log and are introduced during feature encoding. These features are known as engineered features. One potential problem with introducing engineered features is that they might contribute to the loss of interpretability of a regression model~\cite{lipton2018}. However, the features that are engineered in a meaningful way may be aligned with understanding of the business process, in which case, it is likely that they may be interpretable given relevant process knowledge. 

\vspace*{-1.75\baselineskip}
\subsubsection{Analysis~5: Data leakage.} 

We also investigate the predictive models trained for outcome prediction of \textit{bpic2015\_5} event log by using \textit{single\_agg} and \textit{prefix\_agg}, respectively, with XGBoost. Both models have a high accuracy and hence are of interest for deriving interpretations. 

\begin{figure}[h!]
		\vspace*{-.5\baselineskip}
		\resizebox{\textwidth}{!}{\centering\includegraphics{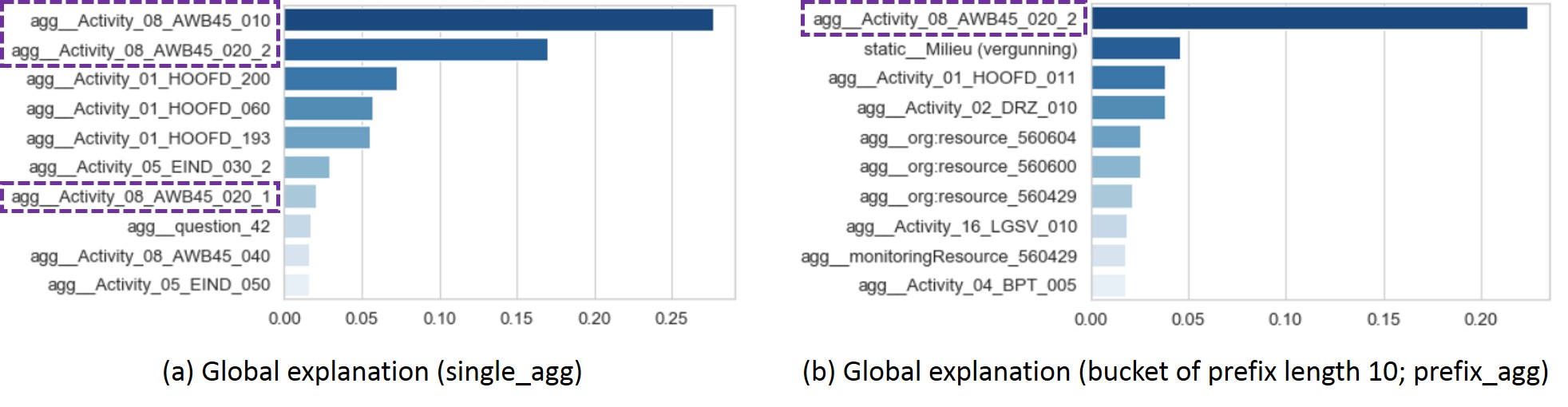}}
    \caption{Interpreting outcome prediction for \textit{bpic2015\_5} event log with XGBoost: (a) global explanation when using \textit{single\_agg} method; and (b) global explanation for bucket of prefix length $10$ when using \textit{prefix\_agg} method.}
\label{fig:bpic2015_agg_single}
\end{figure}

Fig.~\ref{fig:bpic2015_agg_single}(a) depicts the global explanation for the model using \textit{single\_agg} method, which indicates the occurrences of activities \texttt{08\_AWB45\_010}, \texttt{08\_AWB45\_020\_2} and \texttt{08\_AWB45\_020\_1} are three of the important features for outcome prediction. However, the occurrence of \texttt{08\_AWB45\_020\_1} is the outcome to be predicted (as described in Sect.~\ref{subsec:data}). Statistical analysis of the event log shows: i) activity \texttt{08\_AWB45\_020\_2} is executed after \texttt{08\_AWB45\_020\_1} in 68\% of the cases, and ii) activity \texttt{08\_AWB45\_010} occurs at the same time as \texttt{08\_AWB45\_020\_1} in 50\% of the cases. Further analysis also reveals that activity \texttt{01\_HOOFD\_020}\footnote{As described in Sect.~\ref{subsec:data}, activity \texttt{01\_HOOFD\_020} is expected to occur before activity \texttt{08\_AWB45\_020\_1} as part of the temporal rule for labelling process outcome for \textit{bpic2015\_5} event log~\cite{teinemaa2019}.} occurs only once in all cases. All these observations reveal that the predictive model of \textit{single\_agg} with XGBoost exhibits a problem of \textbf{data leakage}~\cite{kaufman2011}, ``where information about the label of prediction that should not legitimately be available is present in the input''. The features that occur along with or after the activity used as the label influence the model predictions. 

Similarly, Fig.~\ref{fig:bpic2015_agg_single}(b) depicts the global explanation of bucket length $10$ when using the model of \textit{prefix\_agg} method with XGBoost, from which we observe activity \texttt{08\_AWB45\_020\_2} as an important feature used for prediction. This also reveals data leakage as \texttt{08\_AWB45\_020\_2} occurs after \texttt{08\_AWB45\_020\_1}. In both scenarios, model explanations along with the knowledge of the business process can be used to identify potential issues with a predictive model.

\vspace*{-\baselineskip}
\section{Discussions} 
\label{subsec:disc}

In this section, we synthesise the findings based on our analyses of interpretations in the previous section. 
The interpretations were derived from selected predictive methods used in the two predictive process monitoring benchmarks~\cite{verenich2019,teinemaa2019}, 
and in summary, they were used to: 
(i) analyse the impact of two different bucketing techniques on a predictive model (Analysis~1); 
(ii) analyse the impact of two different feature encoding techniques on a predictive model (Analysis~2); 
(iii) reveal the impact of one-hot data encoding on a predictive model (Analysis~3); 
(iv) reason the relevance of features identified important to predictions (Analysis~4); and 
(v) identify potential data leakage during a prediction (Analysis~5). 
Below, we discuss several findings, which can be drawn from these analyses, to address the usefulness of model interpretations as well as challenges posed by support to generating model interpretations. 

\paragraph{\it Finding~1: Model interpretations are useful in choosing a suitable prediction method.}
We have observed that while certain bucketing and encoding techniques result in higher accuracy, there is a need to derive interpretations from the model and analyse the features used by the model. 
Model interpretations can be used to review the suitability of different bucketing and encoding mechanisms for predictive models, and also to avoid potential issues (e.g., data leakage) that may incur to predictive models. 
Hence, model interpretations are a valuable input for deciding on a predictive method to use, while so far such a decision is often made by relying on performance measures only. 

\paragraph{\it Finding~2: Model interpretations with domain knowledge enable the understanding of the relevance of features used for predictions.} 
We have observed that model interpretations make it possible to reason about the relevance of features used for predictions, including which certain perspectives of a business process (control-flow, resource, data, time) were used by a predictive model. To gain a deeper understanding of interpretations, the domain knowledge of the relevant business process is also necessary. 

\paragraph{\it Finding~3: Model interpretations can help improve the interpretability of predictive models.} 
As we have also learned from our analyses, the majority of the encoding methods have their advantages and challenges in what concerns the accurate representation of features extracted from event logs capturing business process execution. For example, the aggregation encoding is accurate but abstracts information from the process that could be valuable from an interpretability point of view. This leads to the challenge and dilemma of representing event log data in a way that it is interpretable and accurate. Our analysis also showed that the one-hot-encoding technique generated very sparse feature dimensions, which impacted negatively the explainability of a prediction method. Finally, model interpretations can help reveal engineered features, which are an important input for minimising the incorporation engineered features that may add complexity to the event log or limit the interpretability of a predictive model.

\paragraph{\it Finding 4: Providing business insights from model interpretations.} 
The goal of predictive process monitoring is to provide business users with interesting and useful insights about the underlying business executions. As we have learned from our analyses of model interpretations, relying on performance measures is not adequate to guarantee a good predictive model. It would be useful to a business user if a predictive model can provide additional insights into why a particular prediction was made. 
Providing such an insight is, however, a very challenging task, and remains as an open research question in the scientific community~\cite{rudin2019}.

\section{Conclusions}

In this paper, we have reviewed two existing benchmarks in predictive process monitoring, and presented model interpretations as examples to demonstrate that it is not enough to judge predictive methods by solely relying on their performance measures. Our analyses indicate that accurate models would require deriving interpretations for using the right set of predictive models. Findings drawn from our analyses indicate the need and benefits for using model  interpretations, such as for identifying suitable encoding and bucketing techniques, revealing the importance of features and reasoning the relevance of features used for predictions, detecting potential occurrences of data leakage, etc. Hence, we suggest to incorporate interpretability in addition to the evaluation of predictive models using conventional performance measures (such as accuracy). As a first step into future work, it is important to develop a systematic approach for incorporating model interpretability in predictive process analytics.


\paragraph{\bf Acknowledgement:} We particularly thank the authors of the two process monitoring benchmarks~\cite{teinemaa2019,verenich2019} for the high quality code they released which allowed us to explore model interpretability for predictive process analytics.


\bibliographystyle{splncs}
\bibliography{main}

\end{document}